\documentclass[conference]{IEEEtran}
\IEEEoverridecommandlockouts
\usepackage{cite}
\usepackage{amsmath,amssymb,amsfonts}
\usepackage{algorithmic}
\usepackage{graphicx}
\usepackage{textcomp}
\usepackage{xcolor}
\usepackage{multirow}
\usepackage{array} 
\usepackage{varwidth} 
\usepackage{booktabs} 
\usepackage[caption=false,font=footnotesize]{subfig}
\def\BibTeX{{\rm B\kern-.05em{\sc i\kern-.025em b}\kern-.08em
    T\kern-.1667em\lower.7ex\hbox{E}\kern-.125emX}}

\begin{document}

\title{A Study on the Impact of Data Augmentation for Training Convolutional Neural Networks in the Presence of Noisy Labels \\
{\footnotesize}
\thanks{}
}

\author{\IEEEauthorblockN{Emeson Pereira}
\IEEEauthorblockA{PPGIA, Department of Computing \\
Universidade Federal Rural \\de Pernambuco, Recife, Brazil \\
emesonsantana@gmail.com}
\and
\IEEEauthorblockN{Gustavo Carneiro}
\IEEEauthorblockA{School of Computer Science  \\Australian Institute of Machine Learning\\
University of Adelaide, Adelaide, Australia \\
gustavo.carneiro@adelaide.edu.au}
\and
\IEEEauthorblockN{Filipe R. Cordeiro}
\IEEEauthorblockA{Visual Computing Lab  \\
Universidade Federal Rural \\ de Pernambuco, 
Recife, Brazil \\
filipe.rolim@ufrpe.br}
}




\maketitle

\begin{abstract}
Label noise is common in large real-world datasets, and its presence harms the training process of deep neural networks. 
Although several works have focused on the training strategies to address this problem, there are few studies that evaluate  the  impact of data augmentation as a design choice for training deep neural networks. In this work, we analyse the model robustness when using different data augmentations and their improvement on the training with the presence of noisy labels. 
We evaluate state-of-the-art and classical data augmentation strategies with different levels of synthetic noise for the datasets MNist, CIFAR-10, CIFAR-100, and the real-world dataset Clothing1M. We evaluate the methods using the accuracy metric. Results show that the appropriate selection of data augmentation can drastically improve the model robustness to label noise, increasing up to 177.84\% of relative best test accuracy compared to the baseline with no augmentation, and an increase of up to 6\% in absolute value with the state-of-the-art DivideMix training strategy.
\end{abstract}


\section{Introduction}

Convolutional Neural Networks (CNNs) have been the state-of-the-art for computer vision tasks in the last decade~\cite{li2021survey}. The training of these models requires large amounts of data with precise annotations in order to provide good generalization capabilities~\cite{luo2018does}. However, human or automatic annotations are susceptible to mislabeling due to human failure, challenging tasks, quality of data, etc ~\cite{algan2021image}. Therefore, the presence of label noise is common and often not considered for most of the proposed CNNs and training strategies in literature~\cite{khan2020survey}
. Moreover, most  real-world large datasets available for
training CNNs for computer vision tasks, such as ImageNet~\cite{imagenet}, have the presence of noisy labels.  Zhang et al.~\cite{zhang2021understanding} show that when the dataset is contaminated with label noise, the performance of the model is decreased. For this purpose, different strategies for dealing with labels have been proposed in the last years~\cite{cordeiro2020survey}.

Recent works have shown the importance of using data augmentation along with training strategies to deal with noisy labels. Zhang et al.~\cite{zhang2021understanding} show that the use of regularisation, such as data augmentation, dropout and weight decay helps to improve the generalization performance. They show that the use of simple data augmentation, such as random cropping and random perturbation of brightness helps to improve the results. Nishi et al.~\cite{nishi2021augmentation} show that the use of weak and strong data augmentation can improve training, showing that not only the choice of the data augmentation matters, but also the intensity of the augmentation. Zhang et al. propose the Mixup~\cite{mixup} augmentation, originally proposed for general model training, but which has shown to improve the robustness of the model in the presence of label noise~\cite{dividemix}. 

Most of the recent methods to deal with noisy labels use classical data augmentation methods, such as random cropping, perturbation of brightness, saturation, hue, and contrast, as present in AutoAugment (AutoAug) ~\cite{autoaugment}. Although the recent advances in training strategies and data augmentation methods, there is no consensus on which is the best state-of-the-art data augmentation for dealing with label noise. Moreover, to the best of our knowledge, there is no work analysing the impact of data augmentation design choice in the training of CNNs with noisy labels in the training set. 

This work proposes the analysis of 13 classical and six state-of-the-art (SOTA) data augmentation methods in order to (1) identify the impact on model training when using different data augmentation strategies, and (2) evaluate the best data augmentation method or combination for different levels of noise rate. We perform the analysis for the symmetric, asymmetric and semantic noise over different noise rates. 
The results are evaluated using a vanilla ResNet-18~\cite{resnet} with standard training and also with the SOTA training strategy DivideMix~\cite{dividemix}. 

\section{Related Work}

\begin{figure}[ht!]
\centering
\includegraphics[width=0.7\columnwidth]{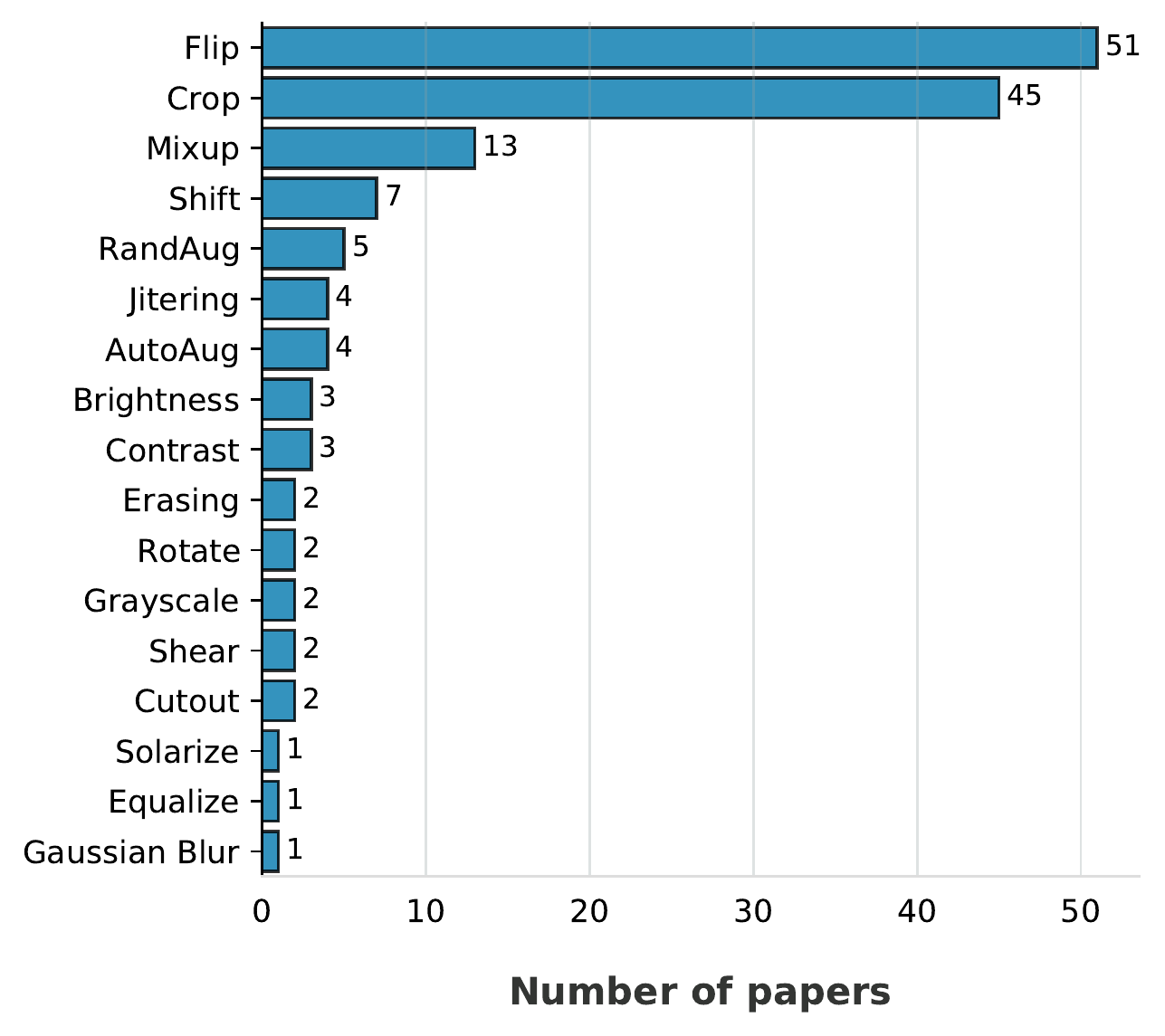} 

\caption{Number of papers related to training with noisy labels that use each type of data augmentation. Sixty-one papers from the last years were evaluated for this analysis.}
\label{fig:papers}
\end{figure}

It is a common practice in modern image classification models to use data augmentation to improve the diversity of the training set and increase model generalization capabilities. Several training strategies have been proposed in the last decade to address the noisy label problem ~\cite{algan2021image}. The main strategies can be classified as robust loss function~\cite{wang2019imae}, sample selection~\cite{han2018co}, label cleansing~\cite{jaehwan2019photometric}, meta-learning~\cite{han2018pumpout} and semi-supervised learning~\cite{dividemix, propmix}. Although the different proposals to train the model, only a few existing data augmentation methods have been explored to address this problem. Figure~\ref{fig:papers} shows a comparison of the most used data augmentation methods present in literature for training  with noisy labels. We evaluated 61 papers related to noisy labels from the last four years and the corresponding data augmentations used. Each paper can use more than one data augmentation method, and we listed the main approaches found in the literature for papers proposed to tackle the label noise problem. 

For the recent strategies to deal with noisy labels, the standard approach is to use classical image processing, such as random cropping and flip, as shown in Figure~\ref{fig:papers}. A few recent approaches~\cite{dividemix,propmix, zhang2020distilling} have started to use SOTA data augmentation methods such as Mixup~\cite{mixup}, CutMix~\cite{cutmix} and RandAugment (RandAug)~\cite{randaugment}.

The Mixup~\cite{mixup} data augmentation proposes a linear combination between samples in order to reduce the empirical vicinal risk. Their approach showed to improve the robustness of the model to label noise and it has been used in training approaches such as DivideMix~\cite{dividemix} and PropMix~\cite{propmix}. The DivideMix approach uses random crop and flipping along with Mixup, whereas PropMix also includes RandAug approach.


Chen et al.~\cite{chen2021robustness} evaluate the training when using no augmentation, standard augmentations (cropping and flip) and strong augmentation (using AutoAug). Their results show an improvement when using strong data augmentation, however their analysis is restricted to one strategy of SOTA data augmentation and one combination of standard augmentation.

Although several recent data augmentation methods have been proposed in the literature, they have not been evaluated in the context of noisy labels. Furthermore, there is no study addressing the analysis of the impact of different data augmentations strategies and the corresponding results for training the model with noisy labels. This work performs this analysis, comparing classical, SOTA and combinations of the data augmentations approaches, evaluating their impact on different datasets and noise rates. In a similar work, Cordeiro et al.~\cite{cordeiro2020survey} show the main strategies to deal with noisy labels in the last years. Although they show the different techniques, there is no experiments related to data augmentation or analysis of data augmentation performance in the label noise context.
Our work aims to show how the choice of data augmentation impacts the training with noisy labels and which strategies can make the model more robust under different noise scenarios.


\section{Methods}

\subsection{Problem Definition}

Lets consider $D=\{(x_1, y_1),...,(x_n,y_n)\}$ as the training set of a classification problem, where $x_i \in \mathcal{X}$ is the $i^{th}$ image and $y_i \in Y$ is a one-hot vector representing the label over $c$ classes. We denote $y \in Y$ as the observed labels (which may or not contain noise). 
  We denote $y_i ^{*}$ as the true label of $x_i$. The distribution of different labels for sample $x$ is denoted by $p(c|x)$, and $\sum_{c=1}^{C} p(c|x) = 1$. 


The overall noise rate is defined by $\eta \in [0,1]$ and $\eta_{jc}$ represents the probability of a class $j$ be flipped to class $c$, as $p(y_i=c|y_i^{*}=j)$.
The main types of noise evaluated in this work are described as follows: symmetric, asymmetric and semantic.

The symmetric noise represents a noise process when a label has equal probability to flip to another class. In the symmetric noise, the true label is also included in the label flipping options, which means that in $\eta_{jc}=\frac{\eta}{C-1}, \forall j \in Y$. Although the symmetric noise is unlikely to represent a realistic scenario for noisy labels, it is the main baseline for noisy label experiments.

The asymmetric noise, as described in \cite{patrini2017making}, is closer to a real-world label noise based on flipping labels between similar classes. 
For asymmetric noise, $\eta_{jc}$ is class conditional. The semantic noise depends on both the classes $j,c\in\mathcal{Y}$ and the image $\mathbf{x}_i$.

\subsection{Data Augmentations}

The data augmentation methods evaluated in this work are divided into two categories: basic and SOTA methods. For the basic augmentations, we evaluated \textit{random crop, horizontal flip, rotation, translation, shear, autocontrast, invert, equalize, posterize, contrast, brightness, sharpness, solarize} and the combination of them. For the SOTA methods we evaluated AutoAug~\cite{autoaugment}, RandAug~\cite{randaugment}, Cutout~\cite{cutout}, Mixup~\cite{mixup}, CutMix~\cite{cutmix}, AugMix~\cite{augmix} and their combination with the previous approaches. 

\begin{figure*}[ht!]
\centering
\subfloat[Original Image]{\includegraphics[width=0.24\columnwidth]{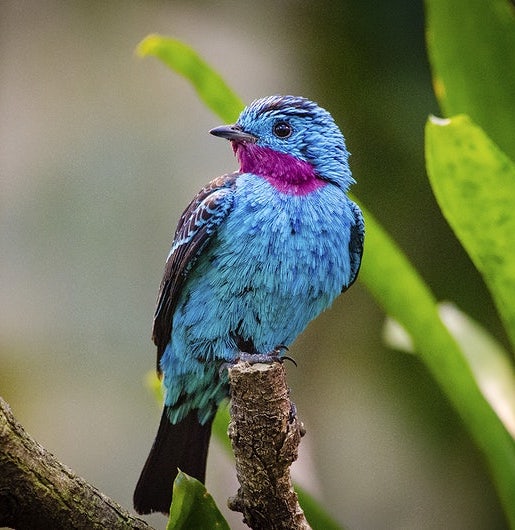} }
\subfloat[Random Crop]{\includegraphics[width=0.24\columnwidth]{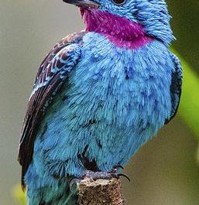} }
\subfloat[Shear]{\includegraphics[width=0.24\columnwidth]{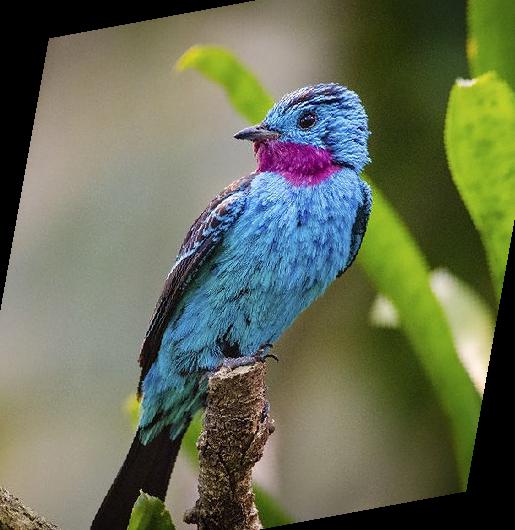} }
\subfloat[Translation]{\includegraphics[width=0.24\columnwidth]{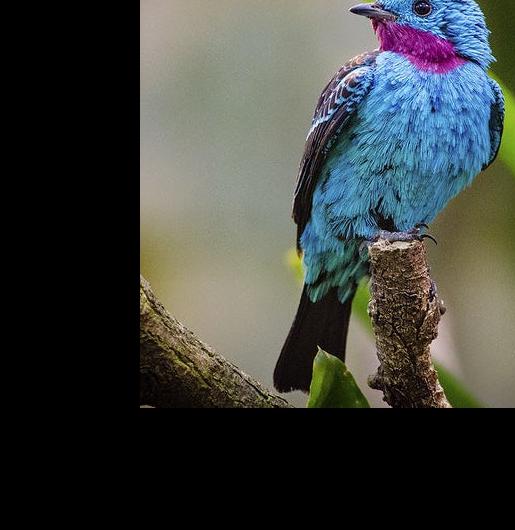} }
\subfloat[Brightness]{\includegraphics[width=0.24\columnwidth]{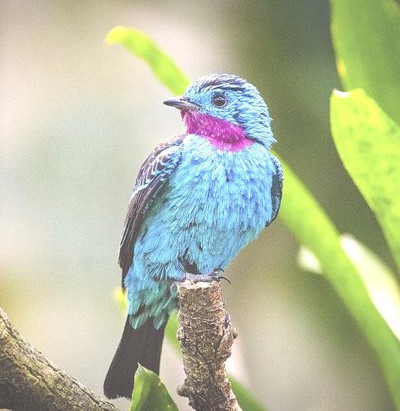} }
\subfloat[Contrast]{\includegraphics[width=0.24\columnwidth]{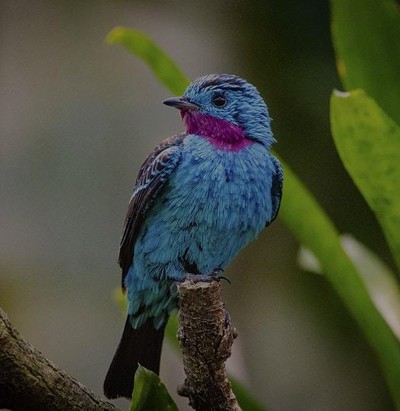} }
\subfloat[Equalize]{\includegraphics[width=0.24\columnwidth]{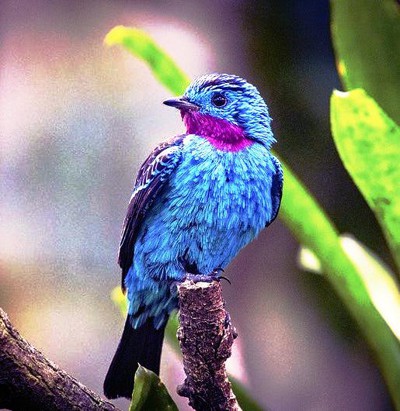} }
\\
\subfloat[Horizontal flip]{\includegraphics[width=0.24\columnwidth]{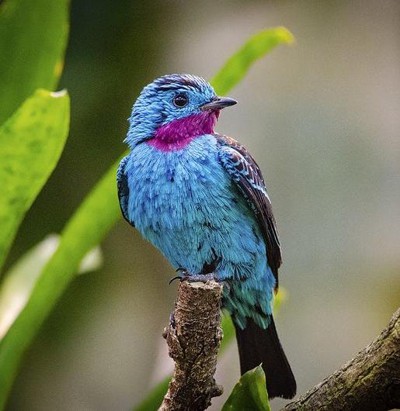} }
\subfloat[Invert]{\includegraphics[width=0.24\columnwidth]{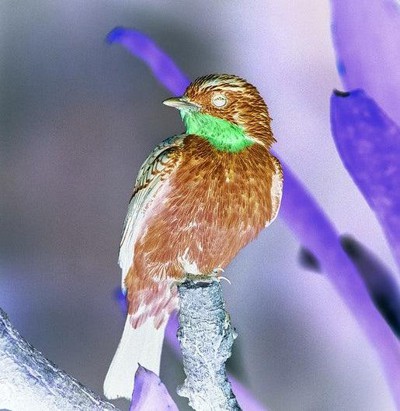} }
\subfloat[Rotate]{\includegraphics[width=0.24\columnwidth]{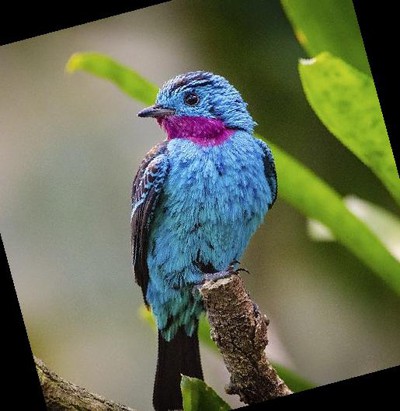} }
\subfloat[Posterize]{\includegraphics[width=0.24\columnwidth]{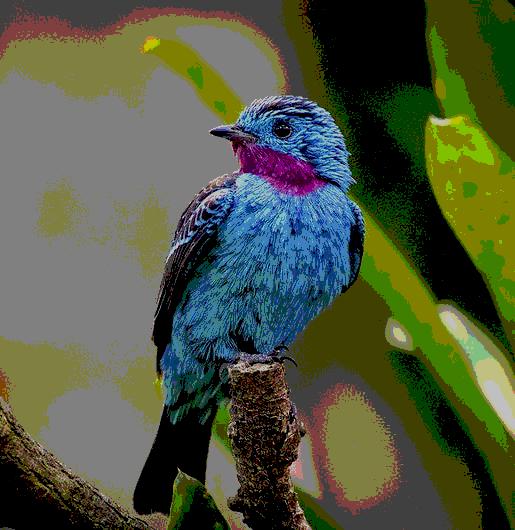} }
\subfloat[Autocontrast]{\includegraphics[width=0.24\columnwidth]{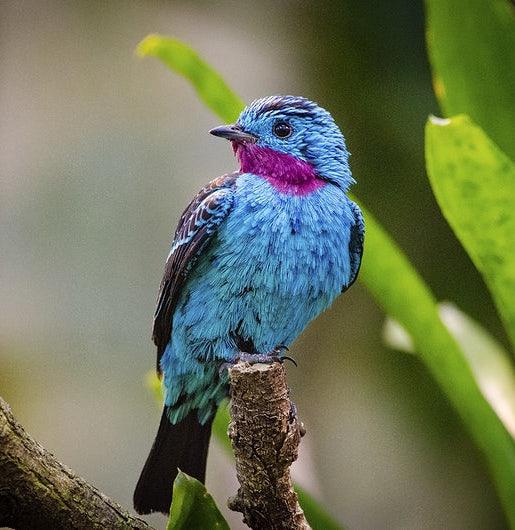} }
\subfloat[Solarize]{\includegraphics[width=0.24\columnwidth]{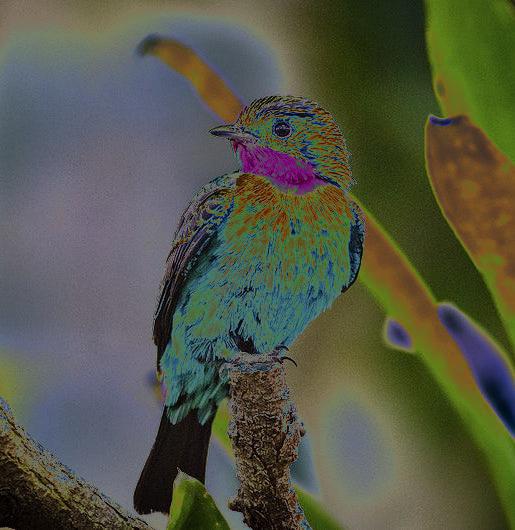} }
\subfloat[Sharpness]{\includegraphics[width=0.24\columnwidth]{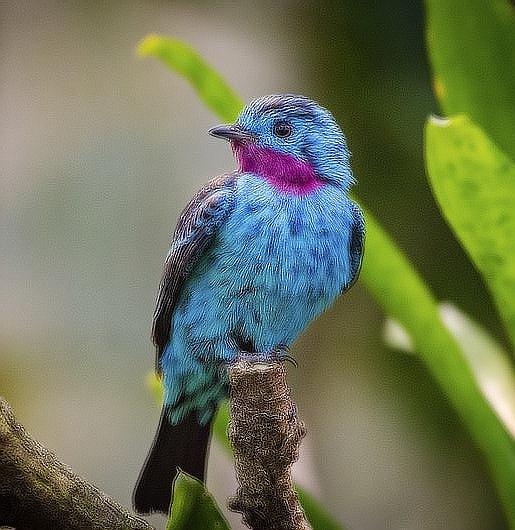} }

\caption{Basic data augmentations based on classical image processing (b)-(n).}
\label{fig:dataug}
\end{figure*}

For the basic augmentations, we used spatial-level and pixel-level transformations on the image, which usually are defined as weak augmentations. We added Table~\ref{tab:classic} describing the effects of each basic augmentation method. Figure~\ref{fig:dataug} shows the different basic data augmentations and their effects on the original image.

\begin{table}[ht]
\caption{Description of the basic augmentations evaluated in this work, separated as spatial-level and pixel-level.}
\label{tab:classic}
\centering
\begin{tabular}{ccV{4.5cm}}
\toprule
Augmentation & Type &  Description  \\
\midrule
random crop & spatial-level & Crop a random part of the input  \\
horizontal flip & spatial-level & Flip the input horizontally around the y-axis \\
rotation & spatial-level & Rotate the input by an angle selected randomly from the uniform distribution \\
translation & spatial-level & Translate the input by swapping rows and columns\\
shear & spatial-level & Distort an input along an axis \\
invert & pixel-level & Invert the input image by subtracting pixel values from 255\\
equalize & pixel-level & Equalize the image histogram\\
posterize & pixel-level & Reduce the number of bits for each color channel\\
contrast & pixel-level & Randomly change contrast of the input image\\
autocontrast & pixel-level & Remap pixels per channel maximizing contrast of an image \\
brightness & pixel-level & Randomly change brightness of the input image\\
sharpness & pixel-level& Sharpen the input image and overlays the result with the original image\\
solarize & pixel-level & Invert all pixel values above a threshold \\
\bottomrule

\end{tabular}

\end{table}


The Mixup~\cite{mixup} data augmentation is one of the SOTA augmentations that has been increasingly applied to noisy label scenarios. It linearly combines images and labels using the following equation:
\begin{equation}
\begin{split}
    \tilde{x} = \lambda x_i + (1-\lambda)x_j, \\
    \tilde{y} = \lambda y_i + (1-\lambda) y_j
\end{split}
\end{equation}

\noindent where $x_i$ and $x_j$ are input images and $y_i$ and $y_j$ are the corresponding observed labels.

Cutout~\cite{cutout} is an image augmentation and regularization technique that randomly masks out square regions of input during training. The CutMix~\cite{cutmix} augmentation, instead of removing pixels and filling them with black pixels, replaces the removed regions with a patch from another image. 

AugMix~\cite{augmix}  method consists of mixing the results from augmentation chains or compositions of augmentation operations. For the input image are applied different sequences of augmentations, such as rotation, translation and posterize, and the images are mixed using different weights. This method has not been explored in the context of noisy labels.

\begin{figure}[ht!]
\centering
\subfloat[Original]{\includegraphics[width=0.22\columnwidth]{images/bird_original2.jpeg} }
\subfloat[Cutout]{\includegraphics[width=0.22\columnwidth]{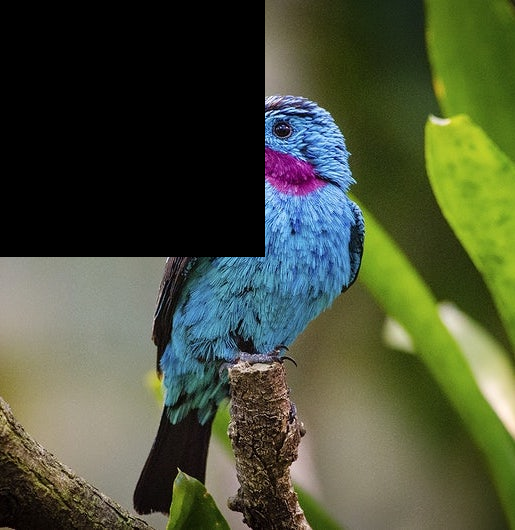} }
\subfloat[Mixup]{\includegraphics[width=0.22\columnwidth]{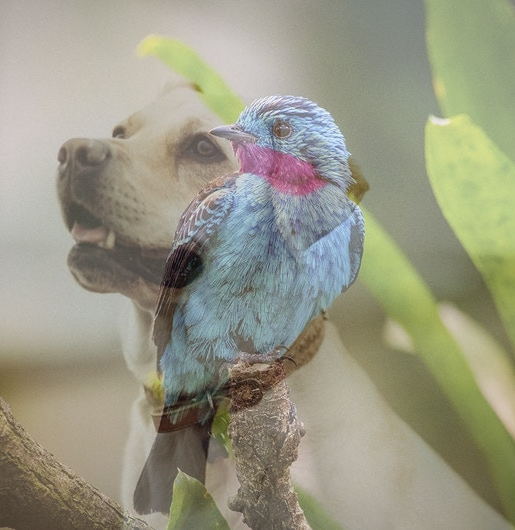} }
\subfloat[CutMix]{\includegraphics[width=0.22\columnwidth]{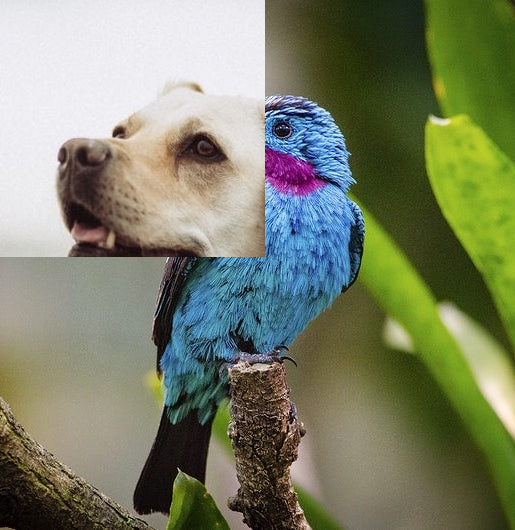} }
\\
\subfloat[AugMix]{\includegraphics[width=0.22\columnwidth]{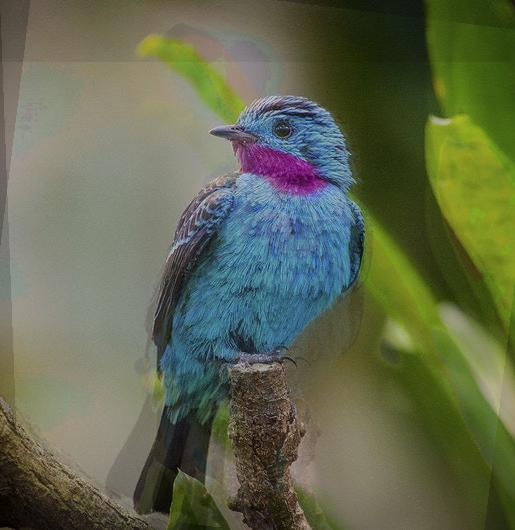} }
\subfloat[AutoAug]{\includegraphics[width=0.22\columnwidth]{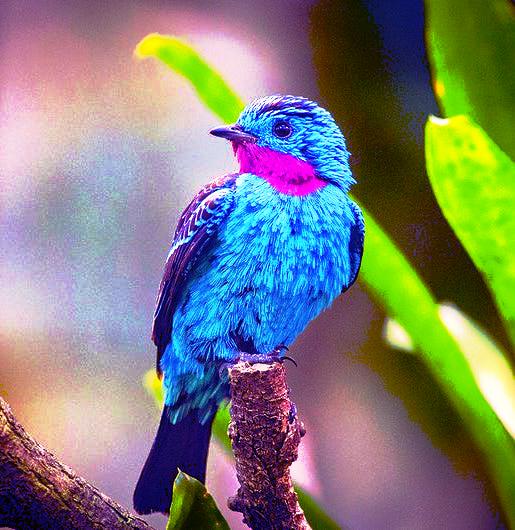} }
\subfloat[RandAug]{\includegraphics[width=0.22\columnwidth]{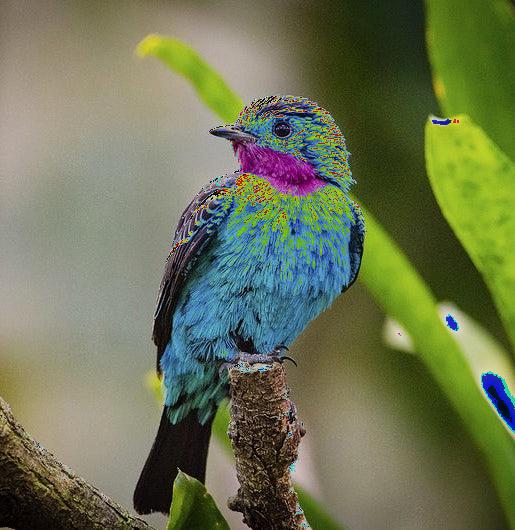} }

\caption{Cutout, Mixup, CutMix, AugMix, AutoAug and RandAug augmentations.}
\label{fig:sota}
\end{figure}

AutoAug~\cite{autoaugment} and RandAug~\cite{randaugment} incorporate augmentation policies during the training, using stronger augmentations optimized for each dataset. A stronger augmentation means a higher intensity in the augmentation, causing a higher perturbation on the image. These augmentations policies include basic transformations such as rotation, shear, invert and contrast.
AutoAug uses reinforcement learning to determine the selection and ordering of a set of augmentation functions
in order to optimize validation loss.
To remove the search phase of AutoAug and therefore reduce training complexity, RandAug drastically reduces the search space for optimal augmentations and uses grid search to determine the optimal set. Figure~\ref{fig:sota} shows Cutout, Mixup, CutMix, AugMix, AutoAug and RandAug augmentations effects on the original image.

\section{Experiments}

\subsection{Datasets}

The experiments were conducted using the datasets MNIST~\cite{mnist}, CIFAR-10~\cite{cifar}, CIFAR-100~\cite{cifar} and Clothing1M~\cite{clothing}.

MNIST consists of handwritten digit images of size 28 $\times$ 28 pixels with a training set of 60,000 samples and a test set of 10,000 samples. For this dataset, we applied synthetic symmetric noise, where each sample has its labels $j$ flipped randomly to another class $c$, with probability $\eta_{jc}$. We evaluated $\eta_{jc} \in \{0.2, 0.4, 0.9\}$, which represents low, medium and high noise scenarios.

CIFAR-10 and CIFAR-100 datasets have 50,000 training and 10,000 testing samples of size 32 $\times$ 32 pixels, where CIFAR has 10 classes and CIFAR-100 has 100 classes. Both datasets have the same amount of samples per class. For this dataset we introduced synthetic symmetric noise, with $\eta_{jc} \in \{0.2, 0.5, 0.8\}$.
We also introduced asymmetric noise for CIFAR-10 and CIFAR-100. The asymmetric noise for CIFAR-10 is produced following the mapping used in~\cite{dividemix}, which maps the classes \emph{truck} $\to$ \emph{automobile}, \emph{bird} $\to$ \emph{plane}, \emph{deer} $\to$ \emph{horse}, with $\eta_{jc} = 0.4 $. The asymmetric noise for CIFAR-100 is produced following the mapping used in~\cite{patrini2017making}, which groups the 100 classes into 20 super-classes containing 5 original classes (e.g., super-class 'Aquatic Mammals' contains 'Beaver', Dolphin', 'Otter', 'Seal', and 'Whale'), and within each super-class the noise flips each class into the next one, circularly.

The Clothing1M dataset contains 1,000,000 clothing images taken from various online shopping sites with noisy labels that reproduce the real world. Clothing1M also contains 14 classes and samples with clean labels, where 50,000 are used for training, 14,000 for validation, and 10,000 for testing. The Clothing1M dataset contains real-world noise, which corresponds to asymmetric and semantic noises.

\subsection{Implementation}

For the implementation of the basic data augmentations, we used the set of parameters following image transformation magnitudes in~\cite{autoaugment}.  
For the SOTA data augmentations AutoAug, RandAug, Cutout, Mixup, CutMix and AugMix, we used the code provided by their official repositories with the default values. For the Clothing1M dataset, data augmentations were applied as in~\cite{dividemix}.

For MNIST, CIFAR-10 and CIFAR-100 we used a ResNet-18 as our baseline model, following~\cite{resnet}. The model for MNIST is trained with stochastic gradient descent (SGD) with momentum of 0.9, weight decay of 0.0005, learning rate 0.001 and batch size of 64. For CIFAR-10 and CIFAR-100, the model is trained with SGD with momentum of 0.9, weight decay of 0.0005, learning rate 0.02 and batch size of 64. We trained the models using 100 and 200 epochs for MNIST and CIFAR datasets, respectively.

In the DivideMix, for CIFAR-10 and CIFAR-100 we used a 18-layer PreAct-ResNet-18 (PRN18)~\cite{he2016identity} as backbone model, as in ~\cite{dividemix}. The model is trained with SGD with momentum of 0.9, weight decay of 0.0005 and batch size of 64. The learning rate is 0.02, which is reduced to 0.002 in the middle of the training~\cite{dividemix}. We trained the model using 300 epochs. 
 
For Clothing1M, we use ResNet-50 as backbone, following ~\cite{dividemix}. In this protocol, a ResNet-50 with ImageNet~\cite{imagenet} pre-trained weights is used. The ResNet-50 is trained for 80 epochs 
 with a batch size of 32, SGD with a learning rate of 0.002 (divided by 10 at epoch 40), momentum of 0.9 and weight decay of 0.0001.

\section{Results}

For MNIST dataset, we evaluate a ResNet-18 with synthetic symmetric noise with noise rate $\eta_{jc} \in \{0.2, 0.5, 0.9\}$ for different types of data augmentation and combinations. We used a vanilla ResNet-18, with no augmentations, as our baseline model. As the MNIST dataset is composed of binary images, we did not use the augmentations \textit{equalize, posterize, solarize, contrast, autocontrast, brightness and solarize}. The results are shown in Table~\ref{tab:res_mnist}. In Table~\ref{tab:res_mnist} we report the best and last values of evaluation on the test set. We denote the random crop augmentation as \textit{rc}.

\begin{table}[ht]
\caption{Results of accuracy using different data augmentations  for MNIST dataset, with $\eta_jc \in \{0.2, 0.5, 0.9\}$. The best values are reported in bold.}
\label{tab:res_mnist}
\centering
\scalebox{0.8}{
\begin{tabular}{c|cc|cc|cc}
\toprule

Method/Noise Ratio & \multicolumn{2}{c|}{20\%} &
\multicolumn{2}{c|}{50\%} & \multicolumn{2}{c}{90\%}  \\

Accuracy & Best & Last & Best & Last & Best & Last \\
\midrule

baseline & 98.31 & 96.70 & 97.16 & 79.21 & 72.96 & 22.04  \\
random crop (rc) & 99.45 & 99.08 & 99.09 & 98.41 & 95.19 & 85.44  \\
random horizontal flip & 97.62 & 95.51 & 96.06 & 68.76 & 72.80 & 20.68  \\
random rotation & 99.22 & 91.90 & 98.72 & 75.72 & 88.89 & 28.77  \\
random translation-x & 99.31 & 91.52 & 98.79 & 72.01 & 88.81 & 27.98  \\
random shear-x & 99.04 & 93.46 & 98.42 & 70.97 & 85.97 & 21.65  \\
random invert & 98.42 & 95.93 & 97.42 & 66.05 & 73.34 & 22.31  \\
random sharpness & 98.41 & 96.39 & 97.18 & 78.73 & 73.39 & 22.82  \\
random crop + translation + shear & 99.57 & 99.46 & \textbf{99.50} & 99.38 & 96.11 & 93.82  \\
all basic transformations & 98.59 & 98.14 & 97.70 & 97.62 & 34.42 & 34.42  \\
\midrule
AutoAug~\cite{autoaugment} & 99.42 & 95.79 & 98.91 & 84.40 & 86.33 & 30.85  \\
RandAug~\cite{randaugment} & 99.48 & 95.13 & 99.06 & 80.11 & 88.88 & 27.50  \\
Cutout~\cite{cutout} & 99.09 & 80.78 & 98.13 & 65.76 & 81.76 & 16.61  \\
CutMix~\cite{cutmix} & 98.36 & 96.05 & 97.20 & 77.11 & 74.90 & 23.16  \\
AugMix~\cite{augmix} & 98.65 & 93.32 & 97.87 & 67.30 & 67.60 & 20.37  \\
Mixup~\cite{mixup} & 98.82 & 96.45 & 97.50 & 79.11 & 73.34 & 21.91  \\
\midrule
AutoAug + random crop & 99.64 & 99.47 & 99.48 & \textbf{99.48} & 96.46 & 94.58  \\
RandAug + random crop & \textbf{99.66} & 99.52 & 99.38 & 99.25 & \textbf{97.10} & \textbf{95.64}  \\
Cutout + random crop & 99.37 & 99.25 & 99.17 & 98.97 & 95.04 & 93.03  \\
Mixup + random crop & 99.47 & 99.31 & 99.17 & 99.05 & 95.40 & 90.68  \\
CutMix + random crop & 99.38 & 99.32 & 99.10 & 98.94 & 95.72 & 94.40  \\
AugMix + random crop & 99.55 & 99.28 & 99.20 & 98.74 & 95.80 & 90.92  \\
\midrule
AutoAug + rc + translation + shear & 99.52 & 99.38 & 99.35 & 99.17 & 92.61 & 91.19  \\
RandAug + rc + translation + shear & 99.58 & 99.50 & 99.40 & 99.23 & 95.06 & 94.74  \\
Cutout + rc + translation + shear & 99.40 & 99.27 & 99.05 & 98.99 & 93.81 & 90.61  \\
Mixup + rc + translation + shear & 99.52 & 99.31 & 99.29 & 99.19 & 93.47 & 91.87  \\
CutMix + rc + translation + shear & 99.43 & 99.23 & 99.20 & 99.16 & 95.08 & 94.34  \\
AugMix + rc + translation + shear & 99.62 & \textbf{99.60} & 99.46 & 99.36 & 96.49 & 94.15  \\

\bottomrule
\end{tabular}
}

\end{table}

\begin{table*}[ht!]
\caption{Results of accuracy using different data augmentations for for CIFAR-10 and CIFAR-100 datasets. The best values are reported in bold.}
\label{tab:res_cifar}
\centering
\scalebox{0.84}{
\begin{tabular}{c|cc|cc|cc|cc|cc|cc|cc|cc}
\toprule
dataset & \multicolumn{8}{c|}{CIFAR-10} & \multicolumn{8}{c}{CIFAR-100} \\
\midrule
Noise type & \multicolumn{6}{c|}{symmetric. } & \multicolumn{2}{c|}{asym.} & \multicolumn{6}{c|}{symmetric} & \multicolumn{2}{c}{asym.} \\
\midrule
Method/Noise Ratio & \multicolumn{2}{c}{20\%} &
\multicolumn{2}{c}{50\%} & \multicolumn{2}{c}{80\%} & \multicolumn{2}{c}{40\%} & \multicolumn{2}{c}{20\%} &
\multicolumn{2}{c}{50\%} & \multicolumn{2}{c}{80\%} & \multicolumn{2}{c}{40\%}  \\

Accuracy & Best & Last & Best & Last & Best & Last & Best & Last & Best & Last & Best & Last & Best & Last & Best & Last \\
\midrule

baseline & 72.60 & 62.31 & 62.96 & 39.67 & 41.50 & 18.25  & 70.30 & 64.64 & 40.27 & 31.42 & 28.94 & 16.31 & 8.27 & 4.92 & 30.67 & 23.60  \\
random crop (rc) & 82.00 & 75.73 & 76.04 & 54.94 & 58.66 & 27.28  & 81.48 & 76.41 & 52.01 & 41.22 & 39.48 & 21.86 & 15.74 & 7.88 & 39.14 & 31.42  \\
random horizontal flip & 78.02 & 68.34 & 68.62 & 44.29 & 48.46 & 18.90 & 76.32 & 65.05 & 45.74 & 34.71 & 34.71 & 16.96 & 17.94 & 4.73 &	35.17 & 28.72 \\
random rotation & 76.32 & 67.82 & 67.76 & 44.60 & 48.93 & 20.24 & 73.99 & 67.83 & 44.79 & 35.21 & 33.75 & 18.01 & 17.12 & 5.86 & 34.54 & 26.59 \\
random translation-x & 78.71 & 70.90 & 69.03 & 45.17 & 50.43 & 20.25 & 77.21 & 69.22 & 45.08 & 35.59 & 34.26 & 19.25 & 18.43 & 5.39 & 33.84 & 30.91 \\
random shear-x & 75.50 & 66.87 & 63.42 & 41.89 & 43.56 & 20.79 & 73.03 & 65.94 & 41.96 & 31.88 & 32.95 & 16.06 & 15.58 & 4.97 & 31.71 & 24.93 \\
random posterize & 72.54 & 63.46 & 62.94 & 39.42 & 44.32 & 18.17 & 69.58 & 65.18 & 40.52 & 30.25 & 30.16 & 14.77 & 13.54 & 4.03 & 30.36 & 23.47 \\
random solarize & 73.48 & 63.18 & 62.80 & 38.15 & 44.87 & 18.54 & 71.76 & 63.86 & 41.05 & 29.33 & 29.91 & 14.22 & 14.55 & 3.98 & 30.45 & 22.97 \\
rc + translation + shear & 82.66 & 81.74 & 75.38 & 70.46 & 55.14 & 46.62  & 80.69 & 79.40 & 51.48 & 46.69& 42.36 & 40.44 & 16.95 & 16.56 & 39.70 & 35.20 \\
all basic transformations & 64.71 & 61.67 & 51.08 & 47.42 & 10.52 & 10.00 & 61.46 & 55.49 & 31.64 & 28.67 & 18.53 & 16.65 & 2.47 & 1.67 & 27.14 & 21.52 \\
\midrule

AutoAug & 77.52 & 67.70 & 67.87 & 42.31 & 46.71 & 19.62 & 75.46 & 65.52 & 44.89 & 34.89 & 34.28 & 16.35 & 16.66 & 4.53 & 34.60 & 28.12 \\
RandAug & 78.89 & 71.86 & 71.86 & 46.03 & 52.22 & 22.19 & 77.18 & 67.00 & 47.49 & 37.46 & 37.67 & 20.22 & 19.54 & 6.42 & 35.87 & 29.29 \\
Cutout & 75.86 & 67.13 & 67.34 & 43.76 & 49.56 & 21.38 & 74.64 & 66.39 & 41.87 & 32.83 & 32.29 & 15.88 & 16.38 & 4.63 & 31.44 & 25.83 \\
CutMix & 79.67 & 74.59 & 70.58 & 51.75 & 50.24 & 22.05 & 76.51 & 68.46 & 50.18 & 42.23 & 38.22 & 23.58 & 18.62 & 6.34 & 39.75 & 31.57 \\
AugMix & 75.38 & 66.68 & 65.53 & 43.61 & 46.10 & 20.30 & 72.30 & 66.21 & 41.38 & 34.02 & 31.12 & 17.13 & 15.07 & 4.85 & 31.75 & 24.34 \\
Mixup & 74.40 & 66.11 & 65.10 & 42.66 & 44.26 & 20.47 & 70.33 & 64.89 & 43.54 & 33.39 & 32.85 & 17.06 & 15.18 & 5.13 & 33.90 & 23.82 \\
\midrule

AutoAug + random crop & 85.05 & 83.56 & 78.84 & 75.88 & 64.10 & 56.76  & 84.16 & 79.57 & 56.13 & 51.73 & 45.26 & 36.97 & 20.39 & 17.09 & 43.40 & 42.26  \\
RandAug + random crop & 84.75 & 82.41 & 77.92 & 72.86 & 62.31 & 49.05  & 82.78 & 79.57 & 55.15 & 45.95 & 43.22 & 25.16 & 19.19 & 11.13 & 41.83 & 33.07 \\
Cutout + random crop & 84.61 & 83.29 & 78.93 & 75.40 & 60.42 & 45.34  & 83.77 & 79.76 & 51.91 & 41.82 & 40.11 & 23.33 & 16.27 & 9.02  & 39.07 & 30.85 \\
Mixup + random crop & 84.53 & 81.88 & 77.15 & 73.93 & 61.64 & 49.08  & 82.53 & 77.79 & 55.64 & 49.30 & 45.10 & 33.45 & 18.56 & 12.23 & 43.46 & 37.61 \\
CutMix + random crop & 85.19 & 82.94 & 78.29 & 71.03 & 63.37 & 50.05  & 83.33 & 77.53 & 56.61 & 52.16 & 46.04 & 32.85 & 19.29 & 12.63 & 48.88 & 43.88 \\
AugMix + random crop & 83.31 & 80.58 & 77.45 & 63.98 & 59.47 & 34.44  & 81.61 & 75.65 & 52.30 & 43.05 & 41.69 & 24.30 & 17.26 & 9.82 & 40.35 & 31.52 \\
AutoAug + rc + translation + shear & 81.15 & 79.74 & 73.40 & 72.59 & 49.28 & 42.70 & 79.53 & 75.25 & 51.34 & 51.34 & 43.40 & 42.16 & 16.31 & 16.31 & 40.84 & 36.05 \\
RandAug + rc + translation + shear & 82.24 & 74.66 & 74.78 & 73.12 & 50.42 & 48.29  & 80.34 & 77.22 & 52.38 & 48.53& 42.95 & 40.99 & 16.59 & 16.55 & 42.73 & 40.43 \\
Cutout + rc + translation + shear & 80.10 & 77.13 & 72.94 & 69.46 & 50.64 & 42.53 & 80.21 & 76.00 & 53.43 & 49.01 & 42.68 & 41.86 & 23.12 & 21.23 & 40.46 & 33.91 \\
Mixup + rc + translation + shear & 80.18 & 76.15 & 72.12 & 66.08 & 49.45 & 44.05 & 78.60 & 72.95 & 48.72 & 45.61 & 41.20 & 37.67 & 17.87 & 15.93 & 41.40 & 39.91 \\
CutMix + rc + translation + shear & 77.85 & 76.76 & 71.59 & 61.34 & 45.79 & 36.60 & 75.74 & 74.62 & 47.34 & 44.82 & 37.73 & 34.15 & 13.38 & 13.18 & 39.80 & 34.06 \\
AugMix + rc + translation + shear & 82.13 & 80.24 & 75.58 & 71.89 & 52.48 & 50.22  & 80.13 & 74.84 & 51.81 & 47.26& 42.94 & 38.36 & 16.22 & 15.71 & 41.51 & 35.34 \\
CutMix + rc + AutoAug & 85.46 & 82.90 & 80.03 & 75.95 & 64.88 & 58.52  & 83.18 & 79.73 & 58.92 & 65.54 & 49.69 & 45.32 & 22.96 & 20.43 & 48.20 & 44.03 \\
Mixup + rc + AutoAug & 85.70 & 83.74 & 80.35 & 75.34 & 65.08 & 60.77  & 82.91 & 79.79 & 59.32 & 54.51 & 48.12 & 48.12 & 20.24 & 17.26 & 49.42 & 45.36 \\
CutMix + Mixup + rc & 84.29 & 82.15 & 77.59 & 74.10 & 62.27 & 53.20  & 82.04 & 72.53 & 56.07 & 52.37 & 42.78 & 35.16 & 17.65 & 16.14 & 48.14 & 46.88 \\
\midrule

DivideMix (w/ Mixup)~\cite{dividemix} & 96.27 & \textbf{96.97} & 94.82 & 94.54 & 93.08 & 92.92  & 93.54 & 91.97 & 77.19 & 76.74 & 74.53 & 74.30 & 54.33 & 54.33  & 60.64 & 53.22\\
DivideMix (w/ Mixup) + AutoAug & 96.28 & 96.14 & 95.32 & 95.11 & 94.18 & 93.96  & 94.41 & \textbf{94.05} & 78.04 & 78.04& 76.53 & 76.06 & 58.98 & 58.52 & 65.44 & \textbf{60.99} \\
DivideMix (w/ CutMix) & 96.42 & 95.96 & 95.25 & 94.82 & 92.00 & 91.46  & 93.22 & 86.97 & 79.76 &  78.84 & 75.64 & 74.11 & 54.42 & 54.15 & 61.74 & 53.79 \\
DivideMix (w/ CutMix) + AutoAug & \textbf{96.91} & 96.46 & \textbf{95.92} & \textbf{95.52} & \textbf{94.39} & \textbf{94.02}  & \textbf{94.74} & 92.34 & \textbf{80.70} & \textbf{80.35}& \textbf{78.68} & \textbf{77.99} & \textbf{60.25} & \textbf{60.03} & \textbf{66.67} & 56.93 \\

\bottomrule
\end{tabular}
}

\end{table*}

From Table~\ref{tab:res_mnist} we can notice that all augmentations improve robustness to noisy labels over the baseline, which has no augmentation. Random crop is the main augmentation to help to deal with noisy labels. The SOTA methods only achieve high performance when combined with random crop. The main reason is because random crop helps to deal with overfitting to label noise, helping the model to learn the target class based on different features of the image. Although the data augmentations individually improve the results, the combination of them, as in \textit{random crop + translation + shear} is more efficient. However, combining all the basic augmentations together, as in \textit{all basic transformations} is less efficient. We also can observe that the SOTA augmentations have better results when combined with random crop. Adding more augmentations, such as in the combination \textit{AutoAug + rc + transition + shear} showed similar results. RandAug augmentation, combined with random crop, showed the best results. 
Compared to the baseline, the choice of a proper data augmentation improved the results up to 33\% on high noise rates.


We also evaluated the augmentations for CIFAR-10 and CIFAR-100 datasets, with symmetric noise $\eta_{jc}\in\{0.2, 0.5, 0.8\}$ and asymmetric noise of 0.4. For these datasets we included all the 13 basic augmentations and SOTA methods and their combinations. We also evaluated the use of SOTA training strategy DivideMix~\cite{dividemix}, using different augmentations. DivideMix standard augmentations use Mixup, random cropping and horizontal flip. We evaluate replacing Mixup for CutMix and adding AutoAug. 
Results are shown in Table~\ref{tab:res_cifar}. We can observe that the addition of basic augmentations to the baseline highly improve the results over different noise rates. As noticed in MNIST dataset, the combination of \textit{rc+translation+shear} showed better results than using individual augmentations. Using all basic transformations together showed to be less efficient. We also noticed that random crop (rc) augmentation is essential to improve to results of SOTA augmentations strategies. From the observed results, the combination of \textit{CutMix+rc+AutoAug} and \textit{Mixup+rc+AutoAug} showed the best combinations of augmentations when added to the baseline. The relative improvement over the baseline test accuracy is up to  61.39\% on asymmetric noise and 177.84\% over high symmetric noise by adding augmentations to the training process. When observing the standard DivideMix strategy, denoted in Table~\ref{tab:res_cifar} as \textit{DivideMix (w/ Mixup)}, the replacement of Mixup for CutMix and the addition of AutoAug improve the results for all the scenarios analysed. The improvement of DivideMix just by changing the augmentation strategy increases the best test accuracy by up to 6\% on CIFAR-100.

We also evaluated the impact of using data augmentation on the real-world dataset Clothing1M. For this dataset, we evaluate the baseline, with no augmentation, and the use of the best strategy observed in the previous datasets (i.e. CutMix+AutoAug). We also compare standard DivideMix with the use of different augmentations.  Results are shown in Table~\ref{tab:res_clothing}. From the observed results in Table~\ref{tab:res_clothing} we can notice an improvement of 1.55\% over the baseline. For Clothing1M, the combination of CutMix and AutoAug showed lower results than standard DivideMix. However, the addition of AutoAug to DivideMix increased the results. 

\begin{table}[ht]
\caption{Results for Clothing1M.}
\label{tab:res_clothing}
\centering
\scalebox{0.9}{
\begin{tabular}{c|c}
\toprule
Method &  Accuracy (\%) \\
\midrule
baseline & 70.30 \\
rc +CutMix + AutoAug & 71.85 \\
\midrule
DivideMix (with Mixup) &  74.32 \\
DivideMix (with Mixup) + AutoAug &  \textbf{75.12}\\
DivideMix (with CutMix) & 72.63 \\
DivideMix (with CutMix) + AutoAug & 69.95 \\

 
 \bottomrule
\end{tabular}
}

\end{table}

With these analyses we show that the choice of data augmentation as a design decision can drastically improve the results when using standard training in the presence of noisy labels. When using SOTA training strategies, the model's performance also can be increased by choosing the augmentation properly. Although the best augmentation strategy seems to be dataset-dependent, we could identify the best augmentation strategies and the importance of choosing them in the training process.

\section{Conclusion and Future Work}

In this work, we evaluated the use of different types of data augmentation in the training of CNNs with the presence of noisy labels. 
The conducted experiments showed that the choice of data augmentation drastically impacts  the training performance. In our analysis, we conclude that the combination of classical approaches and SOTA data augmentations is the best option, and the optimal setup of data augmentations is an important design choice. However, the choice of the best data augmentation is dataset-dependent and needs to be evaluated separatedly.  

For future works we will explore the benefits of using weak and strong augmentations at different stages of the training. We also will investigate new data augmentation strategies and develop some strategies to automatically define the best data augmentation method.


\bibliographystyle{IEEEtran}
\bibliography{main}

\end{document}